\documentclass[11pt,twoside]{article}

\usepackage[utf8]{inputenc}
\usepackage{epsfig,amsfonts,color}
\usepackage{mathrsfs} % provides fancy font for \mathscr
\usepackage{amsmath, bbm, mathabx, amsthm}
\usepackage[font=small]{caption}
\usepackage[normalem]{ulem}
\usepackage{amssymb, palatino, geometry,url}
\usepackage[colorlinks=true,linkcolor=blue,citecolor=blue,urlcolor=blue]{hyperref}
\usepackage{subcaption}
\usepackage{multirow}

\usepackage{graphicx} % Required for inserting images

\usepackage{fancyhdr}
\usepackage{lineno}
\usepackage{float}
\usepackage[section]{placeins}
\usepackage[table]{xcolor}

\title{Physics-Constrained Machine Learning for Chemical Engineering}

\author{Angan Mukherjee and Victor M. Zavala\thanks{Corresponding Author: zavalatejeda@wisc.edu}\\  \\
    {\small Department of Chemical \& Biological Engineering}\\
    {\small \;University of Wisconsin-Madison, 1415 Engineering Drive, Madison, WI 53706, USA}\\
    }
\date{}

\begin{document}

\maketitle

\begin{abstract}
Physics-constrained machine learning (PCML) combines physical models with data-driven approaches to improve reliability, generalizability, and interpretability. Although PCML has shown significant benefits in diverse scientific and engineering domains, technical and intellectual challenges hinder its applicability in complex chemical engineering applications. Key difficulties include determining the amount and type of physical knowledge to embed, designing effective fusion strategies with ML, scaling models to large datasets and simulators, and quantifying predictive uncertainty. This perspective summarizes recent developments and highlights challenges/opportunities in applying PCML to chemical engineering, emphasizing on closed-loop experimental design, real-time dynamics and control, and handling of multi-scale phenomena.
\end{abstract}

\section{Introduction} \label{sec:introduction}

The growing intersection of physics-based modeling and machine learning (ML) models is facilitating the development of more powerful predictive models. Important manifestations of this fusion include physics-constrained neural networks (PCNNs) \cite{zhu2019pcnn} and physics-informed neural networks (PINNs) \cite{raissi2019pinn}. In these {\em hybrid modeling} approaches, the physical laws are directly integrated into the structure and/or into the training workflow of the ML models. This approach provides significant benefits for systems modeling, including better generalization in sparse data regimes and the generation of ML predictions that are physically-consistent \cite{cuomo2022review, meng2025review}. For example, the original PINN framework  \cite{raissi2019pinn}, aimed to train NNs that {\textit{approximately}} satisfy physical equations by adding a regularization term to the loss function that penalizes errors between the NN predictions and the physical equations. Diverse variants of PINNs have been implemented for modeling diverse systems such as photochemical systems, fluid mechanical systems, and heat transfer systems \cite{nilpueng2022heat,sturm2022photochem,raissi2019pinn}.
\\

Researchers have also begun to explore the development of NN models that \textit{exactly} satisfy physical equations/constraints \cite{mukherjee2024mecnn,leug2025dae}. The exact satisfaction of constraints (e.g., mass / energy balances, thermodynamics)  is important in chemical engineering applications  \cite{carranza2022,mukherjee2025metcnn}. A couple of merging paradigms can be identified in these developments such as: sequential projection approaches \cite{chen2024kkthpinn,lastrucci2025enforce,lastrucci2025picard} and simultaneous project approaches  \cite{min2025hardnet,mukherjee2024mcnn,mukherjee2024mecnn,mukherjee2025metcnn}. There exist several variants of these approaches, such as the use of orthogonal projections of linear constraints \cite{chen2024kkthpinn,beucler2021lin}. Recent studies \cite{lastrucci2025enforce,lastrucci2025picard,iftakher2025hardkkt} have proposed strategies for enforcing nonlinear equality constraints. These techniques conduct PCML training by enforcing constraints on the NN outputs \textit {a posteriori} (in a sequential manner). This sequential project approach is flexible and enables the use of existing frameworks for training ML models; however, these might incur high computational costs due to repetitive projection operations, ultimately limiting scalability -- especially for models involving stiff dynamics and highly nonlinear constraints (e.g., thermodynamic relations such as vapor-liquid equilibrium). Another approach is to enforce constraints on the NN outputs by casting the training problem as an optimization problem that simultaneously trains the NN model and enforces constraints; this enables the use of powerful \textit{nonlinear optimizers} such as IPOPT \cite{biegler2006ipopt}, thus facilitating fast superlinear convergence and treatment of complex constraints (equalities and inequalities). However, these approaches are also limited in scalability since NN model components induce dense linear algebra blocks that need to be handled by the optimizer. The inherent limitations of these sequential and simultaneous projection approaches are analogous to those encountered in sequential/simultaneous approaches for solving optimal control problems and problems constrained by differential equations \cite{biegler2007overview}.
\\

The presence of data uncertainty/noise is another important challenge that hinders the development of PCML models, as noise corrupts gradient calculations during learning \cite{nasir2025review}.  Although there approaches for both soft-constrained PCML (e.g., Bayesian PINNs \cite{yang2021bpinn}, fractional PINNs \cite{pang2019fpinn}, conservative PINNs \cite{jagtap2020cpinn}) and hard-constrained PCML \cite{mukherjee2024mcnn,mukherjee2024mecnn,mukherjee2025metcnn} that address  uncertainties (noise, bias) in data, there is a clear need to develop systematic methods for explicitly incorporating  uncertainty of PCML predictions, especially in situations when the available data is incomplete \cite{karniadakis2021review}. Moreover, it is necessary to develop uncertainty quantification (UQ) techniques that can quantify uncertainty of PCML model predictions; UQ capabilities are critical to enable the use of PCML models in closed-loop experimental design tasks and in advanced decision-making frameworks, such as stochastic predictive control.
\\

In this perspective paper, we discuss some of the fundamental open challenges currently persisting in the development of PCML models and in their use in applications of relevance to process systems engineering.
\\

\section{State-of-the-Art Techniques} \label{sec:sota}

\begin{figure}[htbp]
    \centering
    \includegraphics[width=0.99\textwidth]{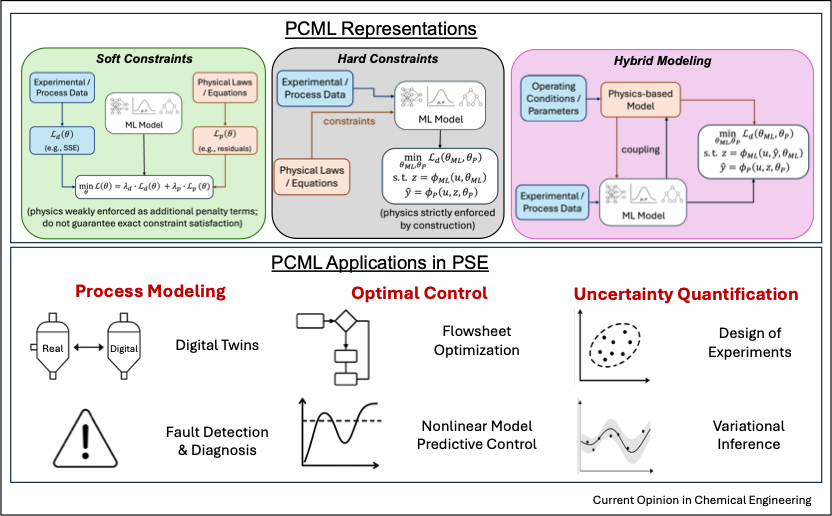}
    \small\caption{Summary of the state-of-the-art PCML models and applications in process systems engineering. Soft-constrained PCML approaches approximately satisfy the physics constraints through penalization of the loss function. Hard-constrained PCML approaches ensure exact satisfaction of physics constraints by imposing constraints directly or sequentially using projection methods. PCML enables advanced analysis and decision-making tasks such as uncertainty quantification, real-time optimization/control, and design of experiments.} 
    \label{fig:sum_fig}
\end{figure}

A visual summary of state-of-the-art PCML paradigms and applications in chemical engineering is presented in Figure \ref{fig:sum_fig}.

\subsection{PCML Problem Statement} \label{sec:prob_statement}

In chemical engineering, we often seek to model complex dynamical systems using input-output data obtained from laboratory and/or process data. We define the available dataset as $\mathcal{D} = {u, y}$, where $u$ are the system inputs and $y$ are the observed outputs. These outputs typically represent a subset of the process states (or a function of such states).

We assume that the \textit {real} system is governed by a set of differential and algebraic equations that encode physical phenomena such as conservation of mass/energy, reaction kinetics, and thermodynamics. In practical applications, we might aim to use available data to build a ML model that serves as a data-driven modeling surrogate of the real system. However, in practical applications, we often have a physics (mechanistic) model that captures aspects of the real system (typically an incomplete/approximate picture). Therefore, our goal is to develop a PCML model that fuses the ML model and the physics model. 

To motivate our discussion, we assume a PCML model structure of the form: 
    \begin{subequations}\label{eq:pcml}
    \begin{gather}
        \quad z = \phi_{ML} (u,\theta_{ML}) \\
        \quad \hat{y} = \phi_P(u,z,\theta_P) 
    \end{gather}
    \end{subequations}
Here, we define $\theta_{\text{ML}}$ as the learnable parameters of the ML model component (represented by $\phi_{ML}$) and we define $\theta_P$ as the learnable parameters of the physics model component (represented by $\phi_{P}$). The PCML model contains intermediate (latent) output variables $z$ that are generated by the ML model and that are fed to the physics model, thus coupling the modeling components as $\phi_{ML}\to \phi_P$ (the physics component corrects the ML predictions). 
\\

\begin{figure}[htbp]
    \centering
    \includegraphics[width=0.99\textwidth]{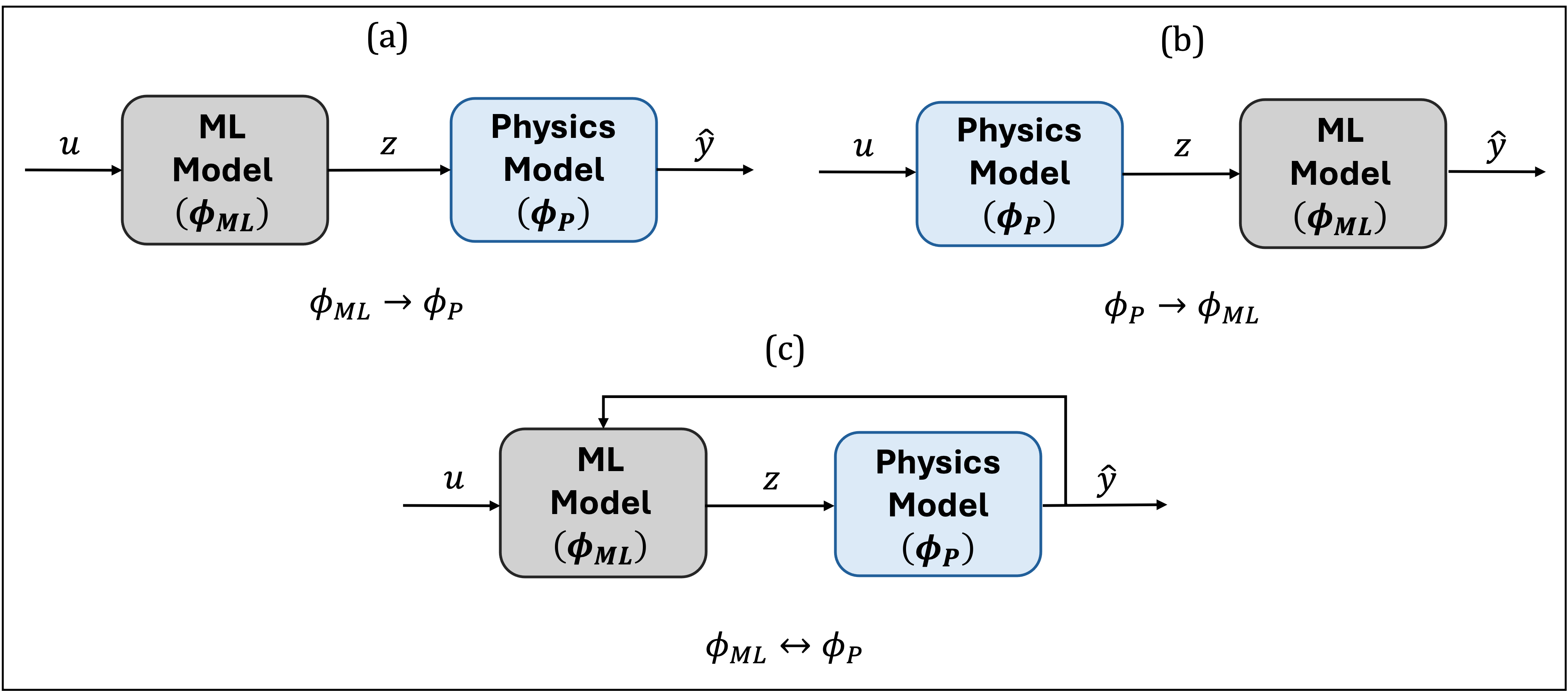}
    \small\caption{Topology of different types of PCML structures. (a) $\phi_{ML}\to \phi_P$: physics component corrects the ML predictions; (b) $\phi_{P}\to \phi_{ML}$: ML component corrects the physics predictions; (c) $\phi_{ML}\leftrightarrow\phi_P$: ML and physics components are linked in a bi-directional manner.} 
    \label{fig:hybrid_fig}
\end{figure}

We highlight that different types of PCML model structures can be envisioned (that couple the ML and physics components in different ways) as shown in Figure \ref{fig:hybrid_fig}; for instance, one can envision a structure in which intermediate outputs of the physics component are fed into the ML component to obtain $\phi_{P}\to \phi_{ML}$ (ML component corrects the physics prediction), or a structure in which the physics and ML components are linked in a bi-directional manner as $\phi_{ML}\leftrightarrow\phi_P$. This generates diverse families of hybrid modeling architectures found in applications. As we will discuss shortly, the nature of the structural coupling of the ML and physics component influences the architecture used to train the PCML model. We note also that the so-called universal differential equation (UDE) model is a type of PCML that follows a structure of the form $\phi_{ML}\to \phi_P$ (the ML component output is propagated through the physics model).
\\

By propagating the intermediate output variables $z$ of the ML component through the physics component, we can express the PCML model using the compact notation:
\begin{subequations}\label{eq:pcml2}
    \begin{align}
         \hat{y} &= \phi_P(u,\phi_{ML} (u,\theta_{ML}),\theta_P) \\
         &=\phi_P(u,\theta_{ML},\theta_P)\\
         &=\phi_P(u,\theta). 
    \end{align}
\end{subequations}
Here, $\theta$ denotes a parameter vector that contains the parameters of the ML and physics components.
\\

The data-fidelity loss, denoted as $\mathcal{L}_d (\theta)$, is typically expressed as a sum of squared errors (SSE) between the measured outputs $y$ and the PCML model predictions $\hat{y}$:

\begin{equation}\label{eq:data_loss}
\mathcal{L}_d(\theta) =  \left( y - \phi_P(u, \theta) \right)^T\left( y - \phi_P(u, \theta) \right)
\end{equation}

The physics loss function, expressed as $\mathcal{L}_p(\theta)$, penalizes the violation of known physical constraints (represented by $\phi_p$) as:

\begin{equation}\label{eq:phy_loss}
\mathcal{L}_p(\theta) = \left( \hat{y}(u, \theta) - \phi_P(u,\theta) \right)^T\left( \hat{y}(u, \theta) - \phi_P(u,\theta) \right)
\end{equation}

\subsection{Taxonomy of Existing PCML Approaches} \label{sec:taxonomy}

Based on existing PCML algorithms discussed in the literature \cite{meng2025review,nasir2025review,bradley2022review}, we can consider the following broad categorization of approaches:

\begin{itemize}
    \item {\textit {Soft-Constrained PCML:} One of the most widely used strategies \cite{raissi2019pinn,karniadakis2021review} enforce NNs to satisfy physics involves the use of soft constraints, which are \textit {weakly} incorporated as additional penalty terms \cite{moya2023daepinn}), as shown in \eqref{eq:sc_pcml}. Here, $\lambda_d$ and $\lambda_p$ are weights between data fidelity and physical consistency. 

    \begin{equation}\label{eq:sc_pcml}
        \min_{\theta} \lambda_d\cdot \mathcal{L}_d(\theta) + \lambda_p\cdot \mathcal{L}_p(\theta) 
    \end{equation}

    Soft-constrained PCML provides implementation flexibility by enforcing physics constraints using first-order learning algorithms; however, this approach does not ensure \textit {exact} conservation of physics. Furthermore, the imbalance in the loss terms can lead to the minimization of a difficult optimization manifold, ultimately leading to slow convergence, sensitivity to hyperparameter tuning, and limited predictive accuracy \cite{wang2022pinnfail}.}
    
    \item {\textit {Hard-Constrained PCML:} In this approach,  ML predictions are required to satisfy physical constraints (typically in the form of equality constraints). This can be done using a couple of paradigms: a sequential projection approach or a simultaneous projection approach \cite{lastrucci2025enforce,mukherjee2024mcnn}, as represented in Figure \ref{fig:proj_fig}. The learning optimization problem that these approaches aim to solve is the following:

    \begin{subequations}\label{eq:hc_pcml}
    \begin{gather}
        \min_{\theta_{ML},\theta_P} \mathcal{L}_d(\theta_{ML},\theta_P) \\
        \text{s.t.} \quad z = \phi_{ML} (u,\theta_{ML}) \\
        \quad \hat{y} = \phi_P(u,z,\theta_P)
    \end{gather}
    \end{subequations}
    
    \begin{figure}[htbp]
    \centering
    \includegraphics[width=0.60\textwidth]{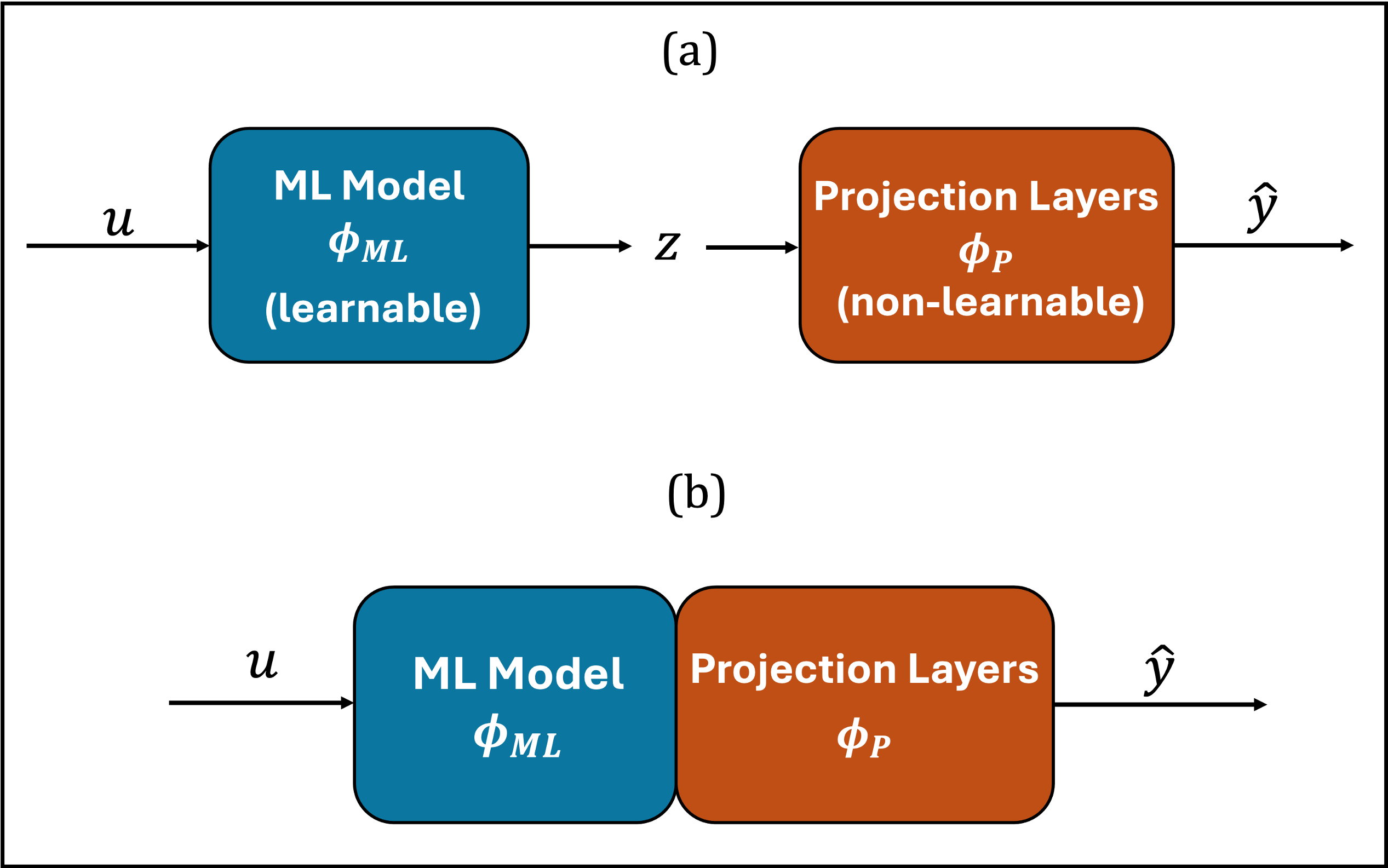}
    \small\caption{Approaches for hard-constrained PCML using (a) sequential and (b) simultaneous projection layers. Sequential projection involves non-trainable layers to recursively project the ML predictions $(z)$ to the physical constraint manifold. Simultaneous projection approaches solve an all-at-once optimization problem and are typically flexible to handle different projection structures.} 
    \label{fig:proj_fig}
\end{figure}
    
    The sequential projection approach exploits the recursive nature of the PCML model in \eqref{eq:hc_pcml} to enforce constraints. Specifically, at each iteration of the learning algorithm, the outputs of the ML model are required to satisfy the physics constraints. This approach enables the use of existing first-order learning algorithms but may suffer of slow convergence, numerical instabilities (e.g., projection might not be possible in early iterations), and high computational costs due to repetitive projection operations.

    In the simultaneous projection approach, the learning optimization problem \eqref{eq:hc_pcml} is solved directly using a constrained nonlinear programming method such as an interior-point method (as implemented in powerful solvers such as Ipopt). This approach can achieve superlinear convergence and enables enforcement of complex constraints and avoidance of instabilities of the sequential approach, but requires solving a more complex problem that that the ML component explicitly as constraints. This also facilitates enforcement of inequality constraints such as variable bounds, which can facilitate the search. Furthermore, it is worth noting that the simultaneous approach can handle general projection structures ($\phi_{ML}\to \phi_{P}$, $\phi_{P}\to \phi_{ML}$, and $\phi_{ML}\leftrightarrow \phi_{P}$), while sequential projection approaches that exist in the literature currently cannot \cite{lastrucci2025enforce,chen2024kkthpinn}. While existing literature also demonstrates that ML components used in PCML models are typically of low complexity (e.g., shallow NNs), the direct handling of constraints by the nonlinear programming solver will lead to scalability limitations \cite{thompson2025pcnode,mukherjee2025metcnn}. This is primarily due to the fact that the size of the optimization problem grows with the number of datasets and because the use of ML components such as NNs creates dense blocks in the linear algebra operations of the nonlinear solver. Here, Schur parallelization and reduced-space linear algebra approaches can be used to partially overcome these limitations \cite{zavala2008paripopt}. 

    }
    
    \item {\textit {General Hybrid Models:} Hybrid/gray-box modeling represents another type of PCML model \cite{sharma2022hybmod}. Unlike hard- and soft-constrained approaches that aim to force ML component predictions to satisfy physics constraints, hybrid models embed partial mechanistic knowledge, such as simplified governing equations or empirical correlations into the learning architecture. This can create complex structures that couple the ML and physics component in the bi-directional form $\phi_{ML}\leftrightarrow \phi_P$. In other words, in these models the PCML predictions are generated by combining the outputs from a physics-based model and of a machine-learned output in a complex (non-sequential) manner  \cite{bangi2022pinnhyb}. For example, the learning optimization problem can take the following form:
        \begin{subequations}\label{eq:hc_pcml2}
    \begin{gather}
        \min_{\theta_{ML},\theta_P} \mathcal{L}_d(\theta_{ML},\theta_P) \\
        \text{s.t.} \quad z=\phi_{ML} (u,\hat{y},\theta_{ML}) \\
        \quad \hat{y} = \phi_P(u,z,\theta_P).
    \end{gather}
    \end{subequations}
    The complex coupling requires of a simultaneous approach for training the model or of more sophisticated learning architectures. 
    }
\end{itemize}

It is important to emphasize that the solution of the learning optimization problem is just a preliminary step towards PCML model building. Specifically, it is necessary to embed cross-validation procedures to ensure that the model generalizes well. Moreover, it is sometimes desirable to quantify the uncertainty of model predictions.

\subsection{Applications of PCML in Chemical Engineering} \label{sec:applications}

PCML has found diverse applications to chemical engineering applications, ranging from heat exchangers \cite{jalili2024pinn} and separators \cite{carranza2022} to polymers \cite{ghaderi2020pinn} and photochemical processes \cite{sturm2022photochem}. The general application areas of PCML can be categorized as surrogate modeling, process control, and UQ.

\begin{figure}[htbp]
    \centering
    \includegraphics[width=0.99\textwidth]{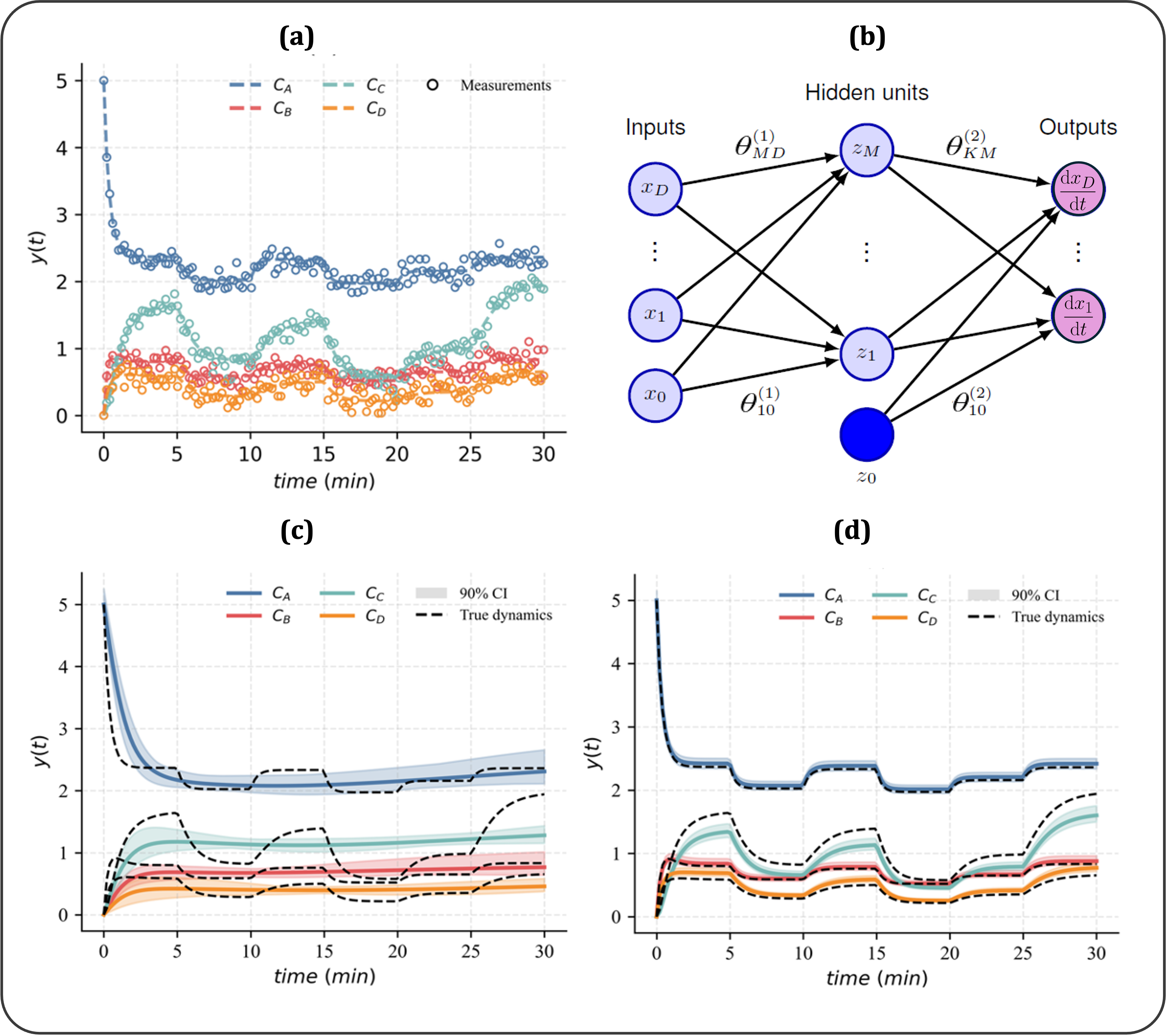}
    \small\caption{ML and PCML modeling of a transient reactor system. (a) Time-series profiles showing true outlet concentrations (dashed lines) and noisy measurements (scatter points) used to train both neural differential model (ML model) and PCML models. (b) Schematic of shallow neural differential model architecture used. (c) Standalone ML predictions with uncertainty bands, trained on noisy data. The model violates physical trends resulting in poor generalization and instability. (d) PCML model predictions using neural differential model with embedded physical constraints. Predictions more closely match true dynamics compared to standalone ML model.}
    \label{fig:appl_fig}
\end{figure}

\begin{itemize}
    \item {\textit {Surrogate Modeling:} In the context of process modeling, PCML addresses the dual challenge of learning interpretable models from limited and/or noisy data while satisfying physical constraints \cite{lastrucci2025enforce,mukherjee2025metcnn}. Soft-constrained frameworks such as PINNs, including their variational and domain-decomposed variants \cite{liu2023varpinn,linyan2024pinndomdecom}, have been used for mesh-free modeling of PDE-governed systems such as adsorption columns, reactive transport processes, and distributed energy networks \cite{subraveti2022pinnadsorption}. The flexibility of PINNs in integrating spatial derivatives and boundary conditions into the learning objective makes them suitable for inverse modeling and system identification. Operator learning models such as DeepONets \cite{wang2021deeponets} and Fourier neural operators \cite{li2024pinnneuraloperator} offer a scalable approach to learning mappings between infinite-dimensional function spaces. These models have found applications in multiscale modeling, surrogate generation for process simulators, and real-time digital twin deployment \cite{wang2021deeponets,li2024pinnneuraloperator,meng2025review, peng2021multiscale}. Hybrid  models are also widely used for systems modeling, thus enabling more accurate and interpretable digital twins, allowing engineers to combine mechanistic insights with data-driven adaptability \cite{bradley2022review}. Several variants of physics-constrained recurrent neural networks (RNNs) have also shown promise in fault detection, soft sensing, and anomaly detection in multivariate time series data \cite{wu2024pimlfault}. 
    } 

    \item {\textit {Real-Time Dynamics and Control:} PCML models can be used as reduced-order surrogates in control applications. Neural ordinary differential equations \cite{thompson2025pcnode} (NODE) and physics-constrained RNNs \cite{zheng2023pinncontrol} have been increasingly adopted for dynamic optimization and control applications because of their ability to represent continuous-time dynamics in a differentiable form. These architectures, when embedded with conservation laws or thermodynamic priors, enable the use of predictive control in energy systems, batch reactors, and bioreactors.  Another emerging direction in PCML-based control is the integration of Koopman operator theory to enable linear representations of nonlinear dynamical systems \cite{rostami2023pinnkoopman}. Recent advances have shown that PCML models can be trained to learn Koopman invariant subspaces while embedding governing constraints, such as energy dissipation, thus showing a strong promise in dynamic optimization tasks where real-time feasibility and interpretability are critical \cite{wang2024pinnkoopman}.
    }
    
    \item {\textit {Uncertainty Quantification:} UQ is a critical issue of PCML models (and of any predictive model), particularly for aiding robust decisions and design of experiments. A notable advantage of hard-constrained PCML is its ability to guide modelers toward selecting the most appropriate set of physical constraints for a given noisy dataset, thus narrowing down the experimental design space \cite{mukherjee2025metcnn}. This feature reduces the manual effort traditionally required to tune or structure dynamic models and opens pathways for rapidly developing reliable surrogate models for optimization and control. On the other hand, soft-constrained PCML methods, particularly Bayesian \cite{yang2021bpinn} and fractional PINNs \cite{pang2019fpinn}, provide more flexible mechanisms for UQ by incorporating probabilistic priors and dropout-based approximations. Advanced parameter estimation techniques such as variational inference (VI) \cite{thompson2025pcnode} provides a promising avenue to train and quantify the uncertainty of PCML models. This is because this approach cast the UQ problem as an optimization problem that can be solved with traditional learning algorithms. Recently, VI has been applied to the development of universal neural differential equation models \cite{nina2025uninode}. Uncertainty characterizations of PCML models has been used to guide the recommendation of experiments \cite{thompson2023bayesiandesign,thompson2025pcnode} (within Bayesian optimization and Bayesian design of experiment frameworks). An important observation is that the use of physics helps narrow down uncertainty of model predictions and helps obtain more consistent uncertainty characterizations. Figure \ref{fig:appl_fig} compares the predictions and uncertainty of a standalone ML model (a shallow neural differential equation model) and PCML model for a reactor example. Here, we use VI to quantify the uncertainty of both models. It is clear that the pure ML model is not capable of adequately capturing the dynamic trends of the data and leads to larger uncertainty bands than the PCML model.
    } 
    
\end{itemize}

\section{Challenges and Future Outlook} \label{sec:challenge & oppor}

Although PCML has shown to improve the accuracy, generalizability, and interpretability of ML models, there are important challenges that need to be addressed to expand the scope of applications.

\begin{itemize}

    \item {\textit {Model structure and Identifiability:} In PCML applications, the available data are typically \textit {limited, noisy, or incomplete}, and governing physical models are often idealized simplifications of the real world. This can create ill-posed learning problems that are difficult to solve and generalize \cite{wang2022pinnfail}. Parameter identifiability becomes particularly challenging when nonlinear interactions occur between learned components and embedded physics, or when multiple physical laws (e.g., mass and energy conservation) must be enforced simultaneously. In practice, constraint violations arise in stiff nonlinear differential equations. These challenges leave open questions: \textit {Which} physical constraints are most important to satisfy? \textit {What} level of physical fidelity is sufficient for reliability? \textit{How} to quantify the information content provided by a physical component?
    Future research opportunities also include automated discovery of physical constraints from data and to explore PCML frameworks that can adaptively select/prioritize constraints during training in order to balance physical fidelity and generalizability/uncertainty. 
    } 

    \item {\textit {UQ and Closed-Loop Design of Experiments:} UQ methods need to produce  uncertainty bounds that respect the embedded physical constraints. In other words, uncertainty estimates need to be physically consistent and this is particularly relevant when designing experiments and in navigating large experimental design spaces. It is necessary to develop Bayesian PCML formulations (such as VI) that can accomplish this. The use of UQ to guide data collection in experimental design procedures can help generate highly information data that refines PCML models and that help discriminate across different PCML architectures (e.g., that embed different types of physical fidelity). This is important, because it is not obvious how to best fuse ML and physical components in PCML \cite{thompson2025pcnode} (this is a practice that remains highly manual).

    }

    \item {\textit {Benchmarking Tools:} A bottleneck in PCML research is the absence of standardized benchmarking protocols, datasets, and diagnostic metrics that capture real settings. In contrast to fields such as computer vision or natural language processing, PCML test problems remain fragmented, often tailored to specific domains such as canonical PDEs, simplified reactors, or synthetic datasets. Such test cases rarely capture the complexities of real-world process systems, which may involve multiple time and length scales, noisy measurements, and interacting subsystems. Existing tools like \texttt{DeepXDE} library \cite{lu2021deepxde} provide valuable frameworks for training PINNs and operator-learning models but remain focused on simple, well-posed PDEs. Similarly, domain-specific tools like \texttt{NeuroMANCER} \cite{drgona2024neuromancer} offer better support for control-oriented PCML applications but are not yet equipped for general-purpose hard-constrained formulations across multiple domains. This gap has led many researchers to adopt ad-hoc implementations, which come at the expense of ease of use, reproducibility, and accessibility \cite{mukherjee2024mcnn,mukherjee2024mecnn}.
    \\

    Future research can largely benefit from developing a domain-agnostic PCML benchmarking suite featuring: (a) reference implementations for soft- and hard-constrained models, (b) curated datasets of varying fidelity and noise levels, and (c) diagnostics for physical violation, generalization, and UQ. Such resources would facilitate rigorous evaluation, fair comparison, and faster transition from isolated case studies to deployable methodologies. However, in practical implementations, several fundamental research questions remain. For example, what design principles can guide the creation of benchmark problems that capture the full spectrum of PCML challenges, i.e., from idealized equations to complex, noisy, multiscale systems? Similarly, how can benchmark metrics be defined to jointly assess predictive accuracy, constraint satisfaction, and computational efficiency?
    }

    \item {\textit {Scalable Computation, Complex Data, and Multiscale Models:} Large-scale PCML faces a couple of intertwined challenges: \textit {computational scalability} and the \textit {effective integration} of heterogeneous, multiscale data sources. Training PCML models often requires solving high-dimensional and highly nonlinear optimization problems, which can become computationally prohibitive. While sequential and simultaneous projection methods can handle certain problems, both of these approaches are limited in the types of PCML models that they can handle. For instance, sequential projection methods can deal with many datasets and complex ML components (e.g., deep NNs) but currently cannot handle complex hybrid structures and suffers of slow (first-order) convergence. On the other hand, simultaneous projection can handle complex hybrid structures but cannot handle complex ML components and many datasets.
    \\

    Future research opportunities include the integration of parallel computation on CPU/GPU architectures and distributed optimization algorithms to accelerate training of simultaneous approaches. On the data side, scalable PCML methods for multi-resolution data fusion, alignment of mixed modalities, and extraction of physically consistent features from noisy or incomplete data will broaden applicability and robustness. 
    \\

    The ability to handle more complex models will enable the creation of PCML model structures that link physical and ML models across multiple scales. For instance, there is a need to link molecular structures to physical properties (using a combination of physical and ML models) and that link physical properties to process properties using process simulators (physical) model (e.g., economics and energy use) \cite{bardow2010solvent,ugo2025fastsolvent} This can be achieved by developing modeling frameworks that can capture diverse types of model structures and fuse them to automate model learning and UQ. 
    
    }

\end{itemize}

\section{Conclusions} \label{sec:conclusions}

In this perspective, we have discussed recent advances in PCML within the context of chemical engineering. We further illustrated their applications to surrogate modeling, process control, and uncertainty quantification, and highlighted their advantages in enhancing reliability, interpretability, and data efficiency. Our analysis underscored that despite these advances, several barriers remain: the difficulty of designing identifiable and well-posed models, the need for uncertainty quantification methods that are physically consistent, the absence of standardized benchmarking tools, and the challenges of scaling PCML to handle large, noisy, and multiscale datasets. We suggest that future work should emphasize automated discovery and prioritization of governing constraints, development of Bayesian and variational formulations for closed-loop experimental design, creation of domain-agnostic benchmarking resources, and exploration of scalable algorithms capable of fusing heterogeneous data streams with high-fidelity simulators. In our opinion, addressing these questions will position PCML as a foundational paradigm for chemical engineering, enabling predictive, interpretable, and physically consistent models that can accelerate progress in experimental design, real-time process control, and discovery of multiscale phenomena.

\section{Acknowledgements} \label{sec:acknowledgements}

We acknowledge support from the National Science Foundation under award CBET-2315963.

\bibliography{ref}
\bibliographystyle{unsrt}

\end{document}